\documentclass[conference]{IEEEtran}

\usepackage{cite}
\usepackage{amsmath}
\usepackage{graphicx}
\usepackage{booktabs}
\usepackage{array}
\usepackage{url}
\usepackage{hyperref}
\usepackage{balance}
\usepackage{fancyhdr}

\title{An Empirical Study of Handcrafted Feature Learning and Convolutional Neural Networks for Facial Expression Recognition}

\author{
\IEEEauthorblockN{Chethiya Galkaduwa}
\IEEEauthorblockA{
Luddy School of Informatics, Computing, and Engineering \\
Indiana University Indianapolis\\
}
}

\begin{document}

\maketitle

\pagestyle{fancy}
\fancyhf{}
\fancyfoot[C]{\thepage}
\renewcommand{\headrulewidth}{0pt}

\begin{abstract}
Facial expression recognition is an important computer vision task with applications in human--computer interaction, mental health monitoring, driver alert systems, and behavioral analysis. While convolutional neural networks (CNNs) dominate modern facial expression recognition, handcrafted feature descriptors such as Histogram of Oriented Gradients (HOG) and Local Binary Patterns (LBP) remain useful classical baselines. This study compares HOG with Support Vector Machine (SVM), LBP with Logistic Regression, and a lightweight CNN across three facial expression datasets: FER-2013, CK+, and KDEF. The results show that CNNs achieve the best overall performance, particularly on more complex data, while HOG performs strongly in controlled environments. LBP performs poorly across all datasets. The study highlights that dataset complexity significantly affects performance and that robust feature learning is essential for real-world facial expression recognition.
\end{abstract}


\section{Introduction}
Facial expression recognition (FER) aims to classify human emotions from facial images. It is widely used in human--computer interaction, mental health monitoring, driver alert systems, surveillance, and behavioral analysis. The task is challenging because facial expressions vary across individuals and are affected by lighting, pose, occlusion, image quality, and dataset conditions.

Modern FER systems are commonly based on deep learning, especially convolutional neural networks (CNNs), because CNNs can automatically learn spatial and hierarchical features from images. However, classical handcrafted feature descriptors such as Histogram of Oriented Gradients (HOG) and Local Binary Patterns (LBP) were widely used before deep learning and remain useful for comparison.

This study investigates whether handcrafted features combined with classical classifiers can still achieve competitive performance compared to a lightweight CNN for facial expression recognition.

The contribution of this work is a systematic empirical evaluation of classical handcrafted descriptors and lightweight deep learning models across datasets with substantially different complexity levels. By analyzing performance degradation from controlled laboratory datasets to unconstrained real-world datasets, this study demonstrates that dataset complexity can have a larger impact on recognition performance than the choice of feature representation alone.

\section{Research Question and Hypotheses}

The primary research question investigated in this study is as follows:

\begin{quote}

Can handcrafted feature descriptors such as Histogram of Oriented Gradients (HOG) and Local Binary Patterns (LBP), when combined with classical machine learning classifiers, achieve competitive performance compared to a lightweight convolutional neural network (CNN) for facial expression recognition?

\end{quote}

The following hypotheses were evaluated:

\begin{itemize}
    \item H1: The CNN-based approach will achieve the highest overall classification accuracy across datasets.
    
    \item H2: The HOG + SVM pipeline will provide competitive performance, particularly on constrained and well-aligned facial expression datasets.
\end{itemize}

\section{Related Work}

Early facial expression recognition systems primarily relied on handcrafted feature extraction techniques such as Histogram of Oriented Gradients (HOG) and Local Binary Patterns (LBP). Dalal and Triggs~\cite{hog} introduced HOG as an edge and gradient-based descriptor that became widely used in object detection and facial analysis tasks. Similarly, Ojala et al.~\cite{lbp} proposed LBP as a texture descriptor capable of encoding local neighborhood intensity patterns.

Traditional facial expression recognition approaches commonly combined handcrafted descriptors with classical machine learning classifiers such as Support Vector Machines (SVMs)~\cite{svm} and Logistic Regression. These approaches achieved strong performance in constrained environments where facial alignment and illumination conditions were controlled.

More recently, deep learning approaches based on Convolutional Neural Networks (CNNs) have become dominant in facial expression recognition research~\cite{deepfer_survey,mollahosseini}. CNNs automatically learn hierarchical feature representations directly from image data and generally outperform handcrafted approaches on large-scale unconstrained datasets such as FER-2013~\cite{fer2013}. Unlike handcrafted descriptors, CNNs jointly optimize feature extraction and classification through end-to-end learning, enabling improved robustness to pose variation, illumination changes, and background noise.

The Extended Cohn-Kanade (CK+) dataset~\cite{ckplus} and the Karolinska Directed Emotional Faces (KDEF) dataset~\cite{kdef} are widely used benchmark datasets for evaluating facial expression recognition systems under controlled conditions.

\section{Datasets}
Three datasets were used to evaluate model performance under different levels of difficulty.

\subsection{FER-2013 (Facial Expression Recognition 2013)}
FER-2013 is an unconstrained real-world facial expression dataset. It contains approximately 35,000 grayscale facial images resized to $48 \times 48$ pixels. The dataset includes seven emotion classes: angry, disgust, fear, happy, neutral, sad, and surprise. Since the images are collected from the internet, FER-2013 contains variations in lighting, pose, occlusion, and image quality.

\begin{figure}[htbp]
\centering
\includegraphics[width=0.48\textwidth]{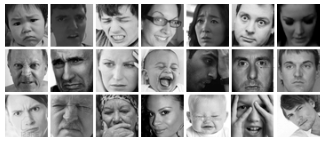}
\caption{Sample images from the FER-2013 dataset showing real-world variation in facial appearance, lighting, pose, and image quality.}
\label{fig:fer2013_samples}
\end{figure}

\subsection{CK+ (Extended Cohn-Kanade Dataset)}

CK+ is a constrained facial expression dataset collected in a controlled laboratory environment. In this study, approximately 980 labeled expression images were used. The dataset contains seven emotion classes, including contempt. The images are generally well aligned and contain clearer facial expressions than FER-2013.

\begin{figure}[htbp]
\centering
\includegraphics[width=0.48\textwidth]{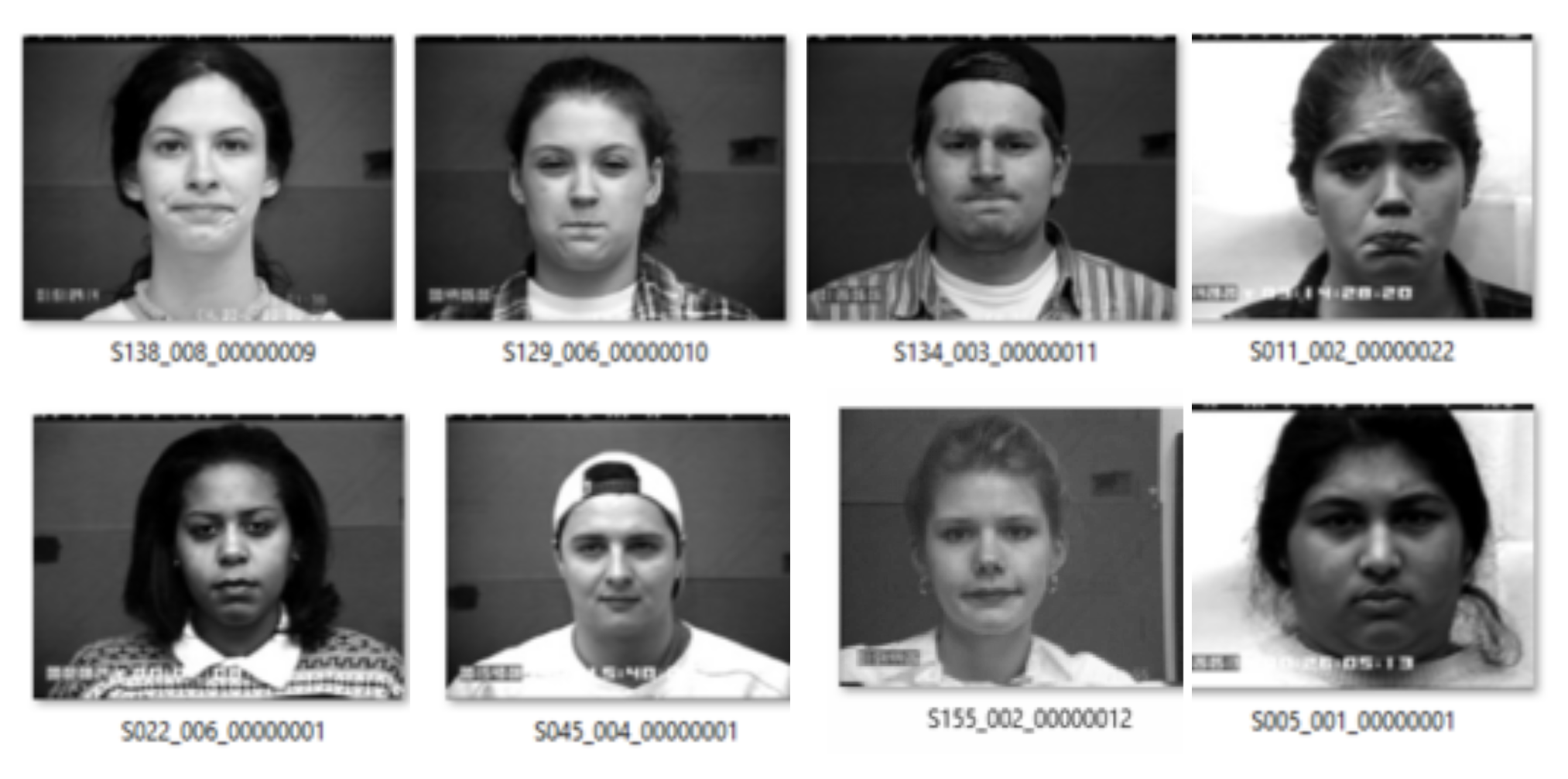}
\caption{Sample images from the CK+ dataset. The images are collected under controlled conditions with aligned faces and clear expression labels.}
\label{fig:ck_samples}
\end{figure}

\subsection{KDEF (Karolinska Directed Emotional Faces Dataset)}
KDEF is a constrained facial expression dataset containing images from 70 subjects and seven emotional expressions. Although the full dataset contains multiple viewing angles, this study used only frontal-facing images, resulting in approximately 980 samples. Compared with CK+, KDEF is more realistic, while still being less complex than FER-2013.

\begin{figure}[htbp]
\centering
\includegraphics[width=0.48\textwidth]{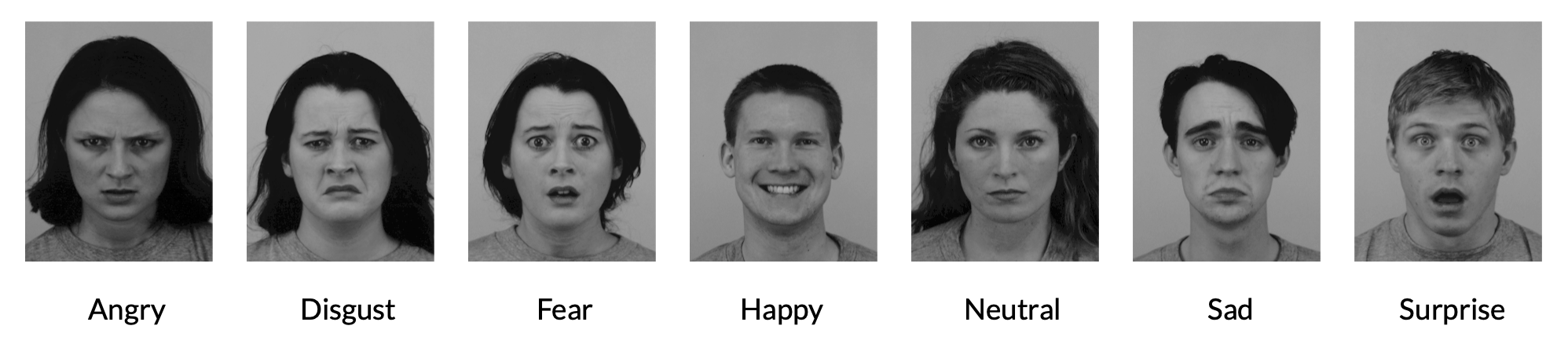}
\caption{Sample frontal images from the KDEF dataset showing seven emotion classes: angry, disgust, fear, happy, neutral, sad, and surprise.}
\label{fig:kdef_samples}
\end{figure}

\subsection{Dataset Complexity Comparison}

The three datasets differ significantly in environmental conditions and classification difficulty. FER-2013 is the most challenging dataset due to large variations in illumination, pose, occlusion, and facial alignment. In contrast, CK+ contains exaggerated and well-aligned facial expressions collected under controlled laboratory conditions. KDEF represents an intermediate level of difficulty, providing more realistic facial appearances than CK+ while remaining less complex than FER-2013.

\begin{table}[htbp]
\centering
\caption{Datasets Used in the Study}
\begin{tabular}{lll}
\toprule
Dataset & Type & Difficulty \\
\midrule
FER-2013 & Unconstrained & High \\
CK+ & Constrained & Low \\
KDEF & Constrained & Medium \\
\bottomrule
\end{tabular}
\label{tab:datasets}
\end{table}

\section{Methodology}
Three approaches were compared: HOG + SVM, LBP + Logistic Regression, and CNN.

\begin{figure}[!t]
\centering
\includegraphics[width=0.48\textwidth]{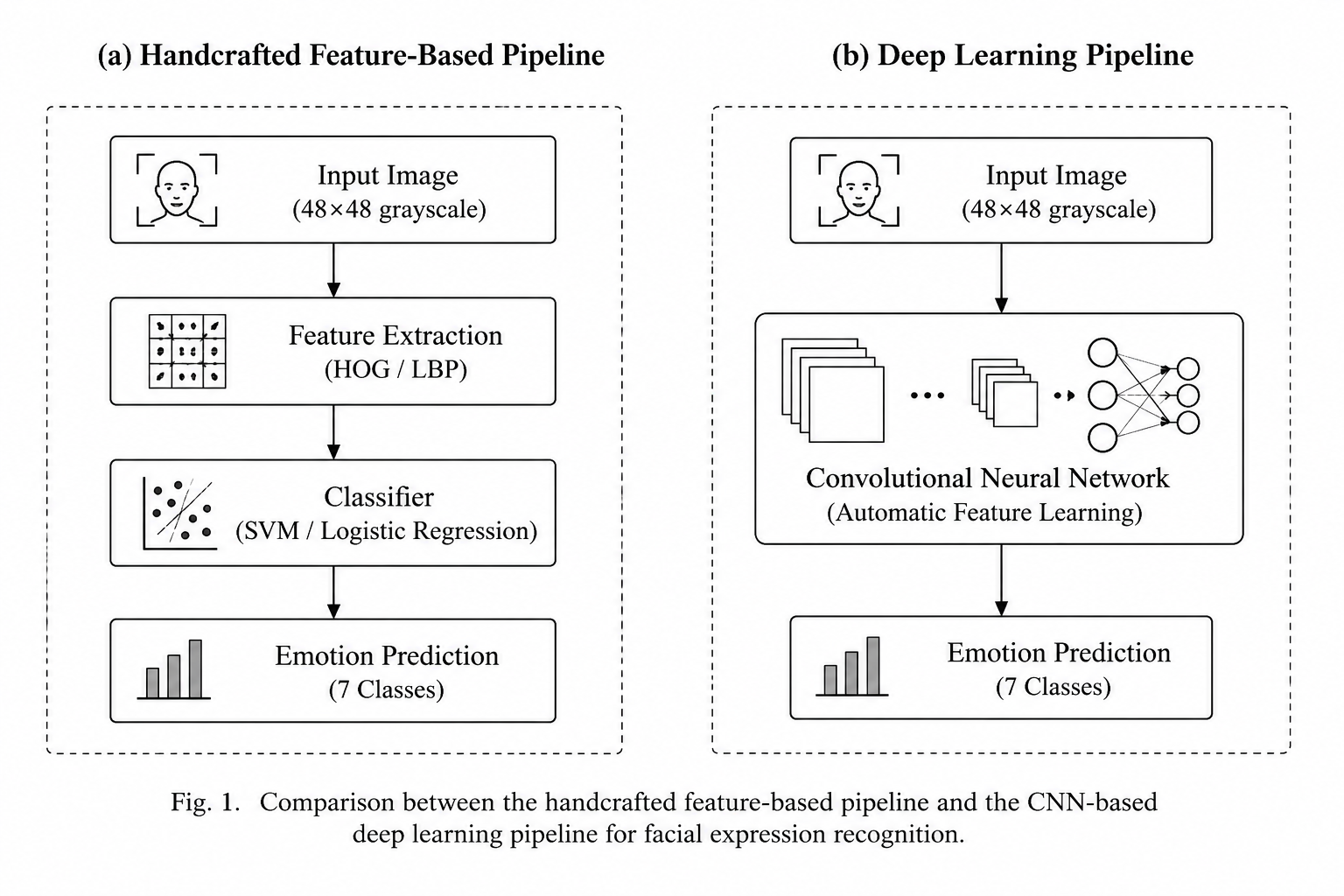}
\caption{Comparison between the handcrafted feature pipeline and the CNN-based deep learning pipeline.}
\label{fig:pipeline}
\end{figure}

As shown in Figure~\ref{fig:pipeline}, the handcrafted pipeline separates feature extraction and classification into independent stages. In contrast, the CNN pipeline performs end-to-end learning, where features are learned automatically during optimization.

\subsection{HOG + SVM}
HOG extracts gradient-based features by computing how pixel intensity changes across the image. It captures edge directions and structural information such as facial contours, mouth shape, and eyebrow movement. After extracting HOG features, a linear SVM classifier was trained to classify the emotion labels.

\begin{figure}[!t]
\centering
\includegraphics[width=0.75\columnwidth]{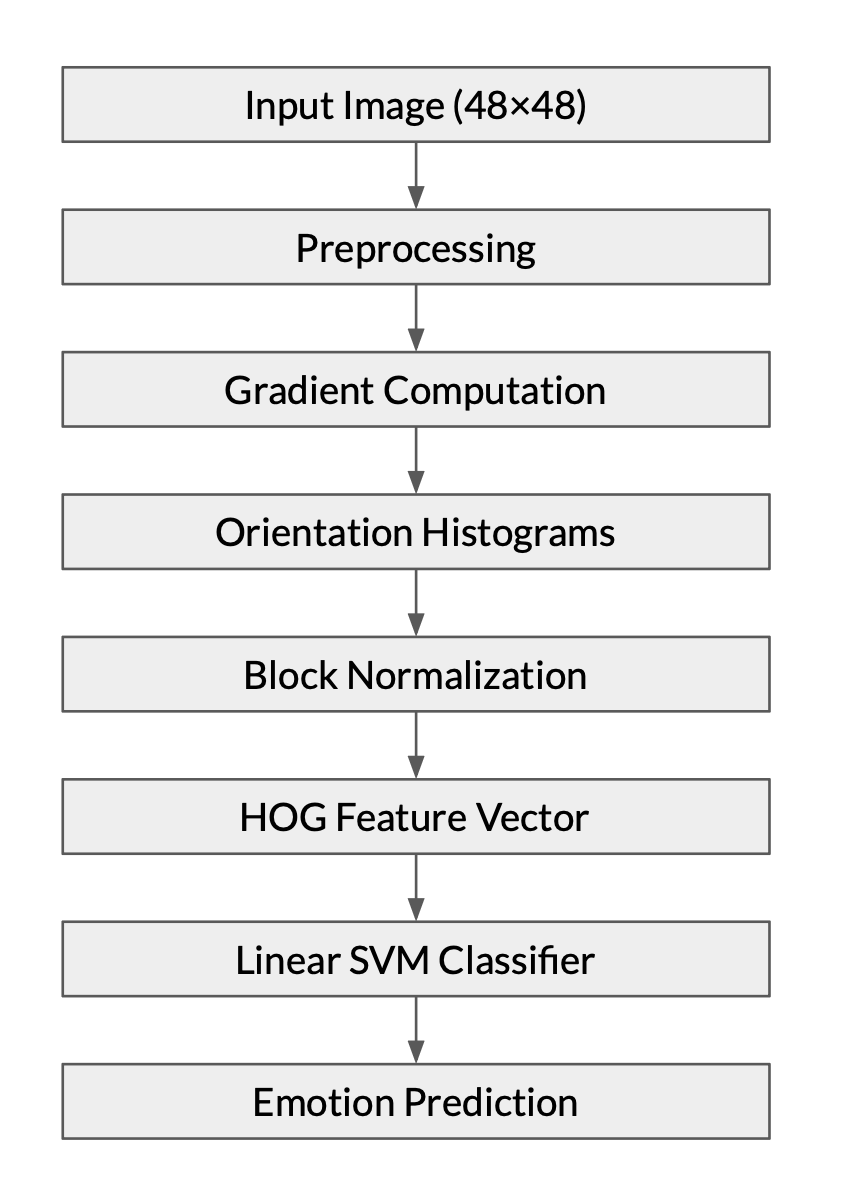}
\caption{HOG + SVM pipeline used for facial expression recognition. The input image is preprocessed, converted into gradient-based HOG features, and classified using a linear SVM.}
\label{fig:hog_pipeline}
\end{figure}

The HOG pipeline in Figure~\ref{fig:hog_pipeline} first converts each $48 \times 48$ facial image into gradient-based descriptors. These descriptors encode edge orientation and facial contour information before classification using a linear SVM.

\subsection{LBP + Logistic Regression}
LBP captures local texture patterns by comparing each pixel with its neighboring pixels. The output is represented as a histogram feature vector. Logistic Regression was then used as the classifier.

\begin{figure}[!t]
\centering
\includegraphics[width=0.75\columnwidth]{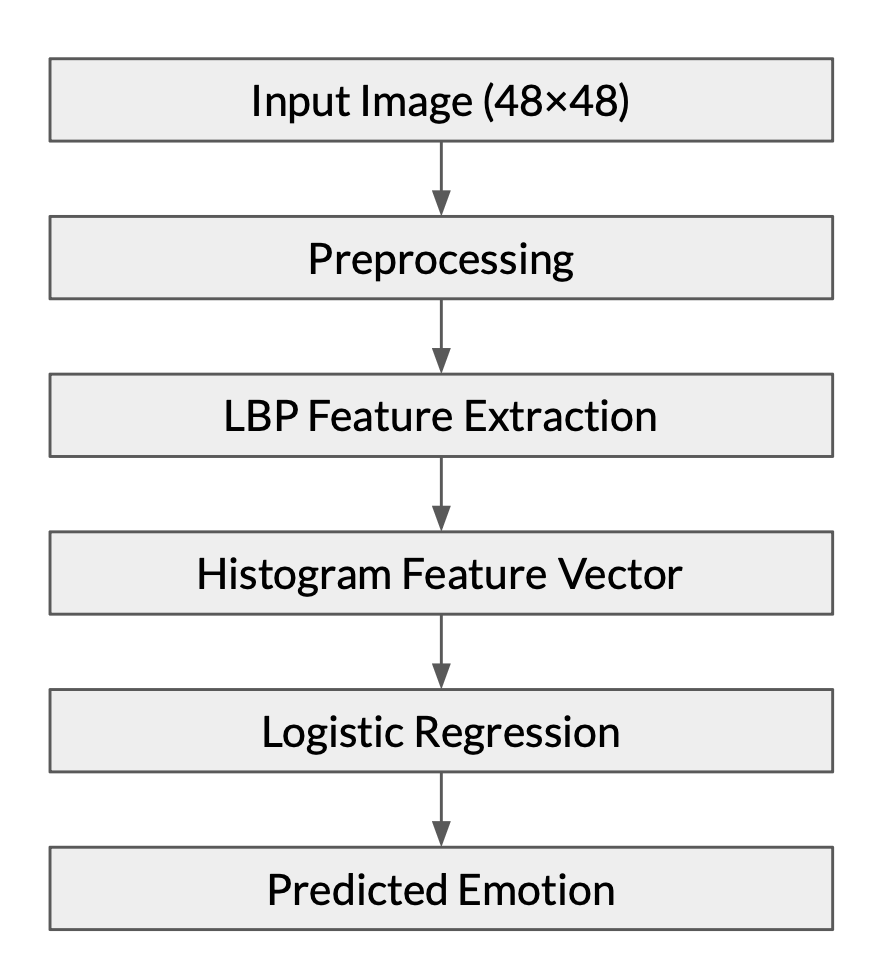}
\caption{LBP + Logistic Regression pipeline used for facial expression recognition. The input image is preprocessed, converted into LBP histogram features, and classified using Logistic Regression.}
\label{fig:lbp_pipeline}
\end{figure}

As shown in Figure~\ref{fig:lbp_pipeline}, the LBP pipeline represents each image using local binary texture patterns. These local patterns are converted into histogram features and then used by Logistic Regression for emotion prediction. LBP focuses on texture rather than facial structure, which can limit its performance in emotion recognition.

\subsection{CNN}
A lightweight CNN was designed to learn features automatically from grayscale facial images. The model used $48 \times 48$ input images and produced predictions for seven emotion classes.

\begin{figure}[!t]
\centering
\includegraphics[width=0.72\columnwidth]{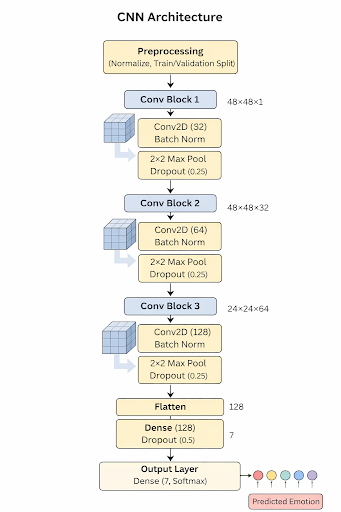}
\caption{Architecture of the lightweight CNN used for facial expression recognition. The network consists of three convolutional blocks followed by fully connected layers and a Softmax output layer.}
\label{fig:cnn_architecture}
\end{figure}

\begin{table}[htbp]
\centering
\caption{CNN Architecture Configuration}
\begin{tabular}{lc}
\toprule
Component & Configuration \\
\midrule
Input Size & $48 \times 48 \times 1$ \\
Conv Layers & 3 \\
Activation Function & ReLU \\
Pooling & MaxPooling \\
Normalization & Batch Normalization \\
Dropout & 0.25 \\
Optimizer & Adam \\
Loss Function & Categorical Cross-Entropy \\
Output Classes & 7 \\
\bottomrule
\end{tabular}
\label{tab:cnn_config}
\end{table}

Figure~\ref{fig:cnn_architecture} shows that the CNN contains three convolutional blocks with Batch Normalization, MaxPooling, and Dropout layers. Feature maps extracted by the convolutional layers are flattened and passed through a dense layer before final emotion classification using a Softmax output layer.

The CNN was implemented using TensorFlow and Keras. ReLU activation functions were used after each convolutional layer, while Softmax activation was used in the final classification layer. The model was trained using the Adam optimizer with categorical cross-entropy loss. Early stopping was applied to reduce overfitting by monitoring validation loss during training.

\section{Experimental Setup}

All facial images were resized to $48 \times 48$ grayscale format before training and evaluation. For FER-2013, the predefined training and testing split provided by the dataset was used. For CK+ and KDEF, custom train-test splits were created because standardized splits were not provided. A subject-wise split strategy was applied to KDEF to ensure that images from the same subject did not appear in both training and testing sets, reducing identity-based bias during evaluation.

For CK+ and KDEF, approximately 80\% of the images were used for training and 20\% for testing. Pixel intensities were normalized to the range $[0,1]$ before feature extraction and CNN training.

The CNN model was implemented using TensorFlow and Keras. Training was performed using the Adam optimizer with categorical cross-entropy loss. Early stopping was applied based on validation loss to reduce overfitting. The handcrafted pipelines used HOG and LBP feature extraction combined with linear SVM and Logistic Regression classifiers, respectively.

Performance was evaluated using classification accuracy, confusion matrices, and classification reports to analyze both overall and class-wise prediction behavior.

\subsection{Evaluation Metrics}

Model performance was primarily evaluated using classification accuracy:

\begin{equation}
Accuracy = \frac{TP + TN}{TP + TN + FP + FN}
\end{equation}

where TP, TN, FP, and FN represent true positives, true negatives, false positives, and false negatives, respectively.

Confusion matrices were also analyzed to evaluate class-wise prediction behavior and inter-class confusion patterns.

\section{Results}

\subsection{HOG + SVM Results}

\begin{table}[htbp]
\centering
\caption{HOG + SVM Accuracy Across Datasets}
\begin{tabular}{cc}
\toprule
Dataset & Accuracy (\%) \\
\midrule
FER-2013 & 44.0 \\
KDEF & 58.2 \\
CK+ & 98.4 \\
\bottomrule
\end{tabular}
\label{tab:hog_results}
\end{table}

Table~\ref{tab:hog_results} summarizes the classification accuracy achieved by the HOG + SVM model across the three datasets. The HOG + SVM model achieved the highest performance on the CK+ dataset with an accuracy of 98.4\%. Moderate performance was obtained on KDEF, while performance dropped significantly on FER-2013. Since CK+ is collected in controlled laboratory conditions with aligned facial expressions and minimal background variation, gradient-based handcrafted features extracted by HOG become highly discriminative. In contrast, FER-2013 contains unconstrained real-world images with illumination variation, pose differences, occlusions, and noisy backgrounds, making handcrafted structural descriptors less robust.

\begin{figure}[!t]
\centering
\includegraphics[width=0.82\columnwidth]{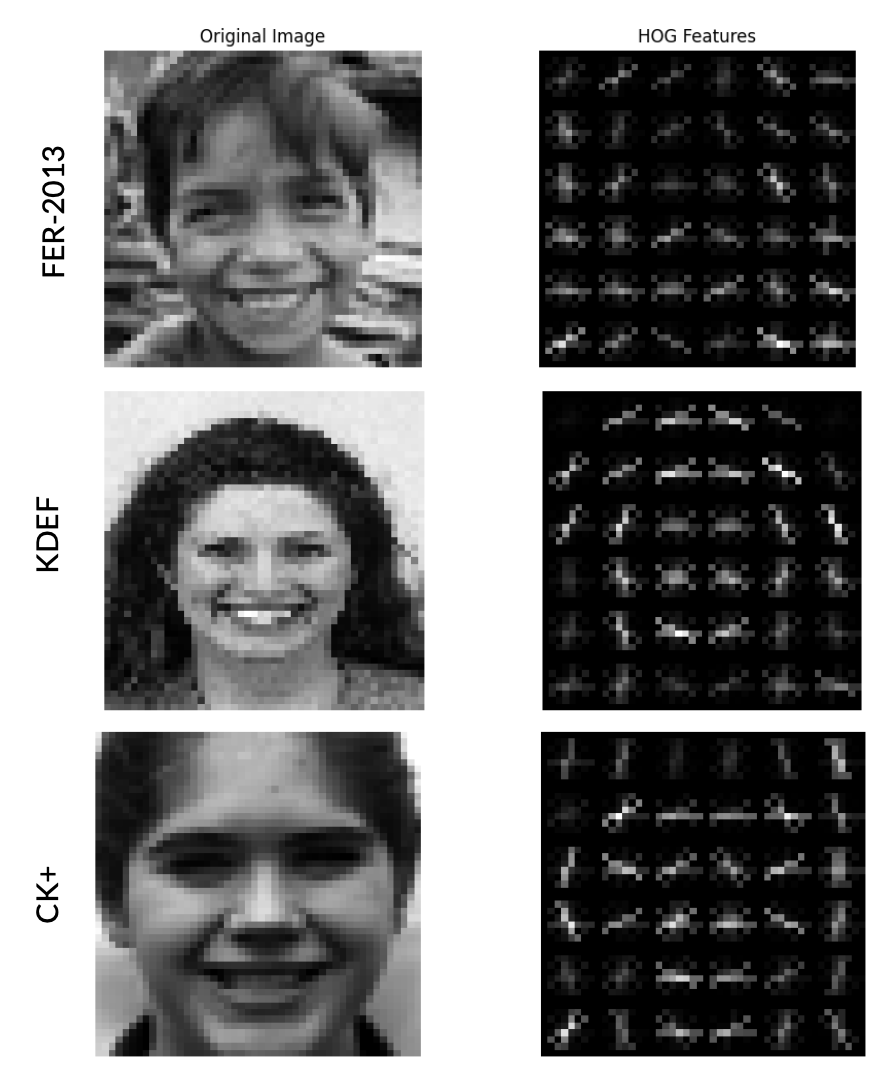}
\caption{Sample HOG feature visualizations from FER-2013, KDEF, and CK+ datasets. HOG primarily captures edge orientation and facial contour information around important facial regions such as the mouth, eyes, and eyebrows.}
\label{fig:hog_visualization}
\end{figure}

Figure~\ref{fig:hog_visualization} illustrates how HOG transforms facial images into gradient orientation representations. Strong responses are observed around facial boundaries and expression-related regions. The visualization demonstrates that HOG mainly focuses on structural edge information rather than detailed texture representations.

\begin{figure*}[!t]
\centering
\includegraphics[width=0.95\textwidth]{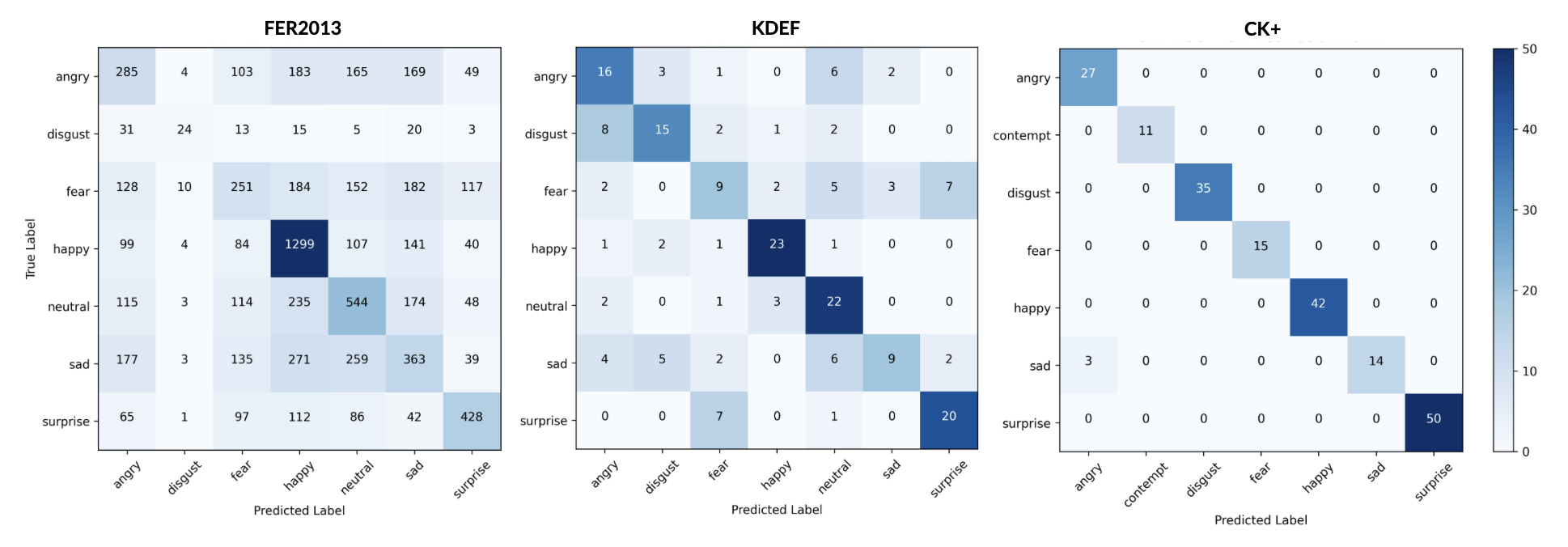}
\caption{Confusion matrices for the HOG + SVM model across FER-2013, KDEF, and CK+ datasets.}
\label{fig:hog_confusion}
\end{figure*}

The confusion matrices shown in Figure~\ref{fig:hog_confusion} further highlight the strengths and weaknesses of the HOG-based approach. On the CK+ dataset, most emotion classes are correctly classified with minimal confusion due to the clean and exaggerated facial expressions available in the dataset. In contrast, FER-2013 exhibits substantial inter-class confusion, particularly between emotions such as fear, sadness, and neutral. This indicates that handcrafted gradient-based descriptors struggle to generalize under unconstrained real-world conditions where facial appearance varies significantly.

The KDEF dataset produced intermediate performance because it contains more realistic facial appearances than CK+ while still maintaining controlled imaging conditions. Although HOG successfully captures major facial contours and expression-related edges, it remains sensitive to variations in illumination, facial alignment, and image quality.

Overall, the HOG + SVM pipeline provides a strong classical baseline for controlled facial expression recognition tasks but demonstrates limited robustness on challenging real-world datasets.

\subsection{LBP + Logistic Regression Results}

\begin{table}[htbp]
\centering
\caption{LBP + Logistic Regression Accuracy Across Datasets}
\begin{tabular}{cc}
\toprule
Dataset & Accuracy (\%) \\
\midrule
FER-2013 & 25.0 \\
KDEF & 14.7 \\
CK+ & 25.0 \\
\bottomrule
\end{tabular}
\label{tab:lbp_results}
\end{table}

The LBP + Logistic Regression model produced significantly lower performance compared to both HOG + SVM and CNN approaches (Table~\ref{tab:lbp_results}). Accuracy remained low across all datasets, indicating that LBP features alone are insufficient for reliable facial expression recognition. The model particularly struggled on KDEF, where the accuracy dropped to 14.7\%.

\begin{figure}[!t]
\centering
\includegraphics[width=0.82\columnwidth]{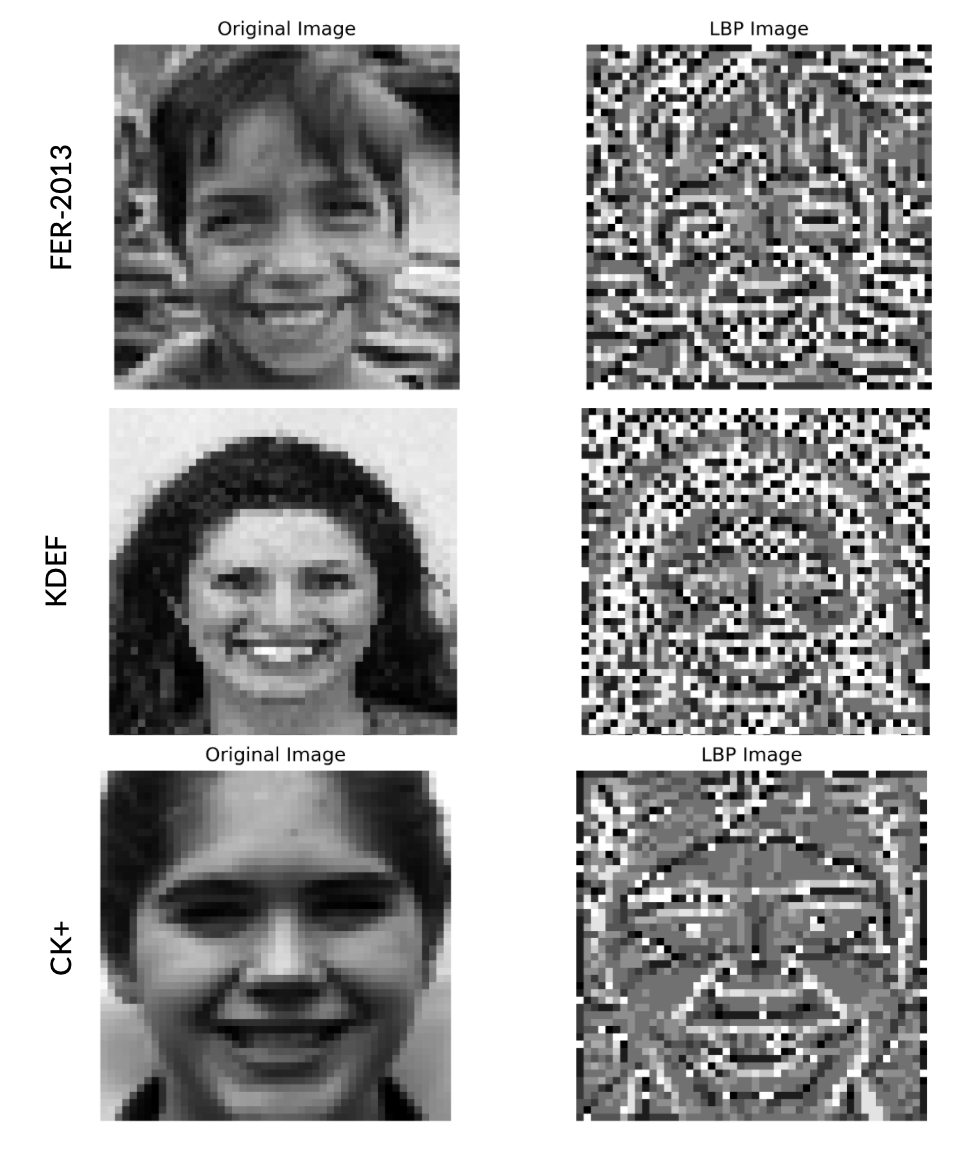}
\caption{Sample LBP feature visualizations from FER-2013, KDEF, and CK+ datasets. LBP captures local binary texture patterns by encoding pixel-level intensity relationships.}
\label{fig:lbp_visualization}
\end{figure}

Figure~\ref{fig:lbp_visualization} illustrates the texture-based representations generated by LBP. Unlike HOG, which focuses on structural edge orientation, LBP mainly captures local pixel intensity transitions and micro-texture information. Although some facial regions remain visible, the resulting representations appear noisy and fragmented, particularly for unconstrained datasets such as FER-2013.

\begin{figure*}[!t]
\centering
\includegraphics[width=0.95\textwidth]{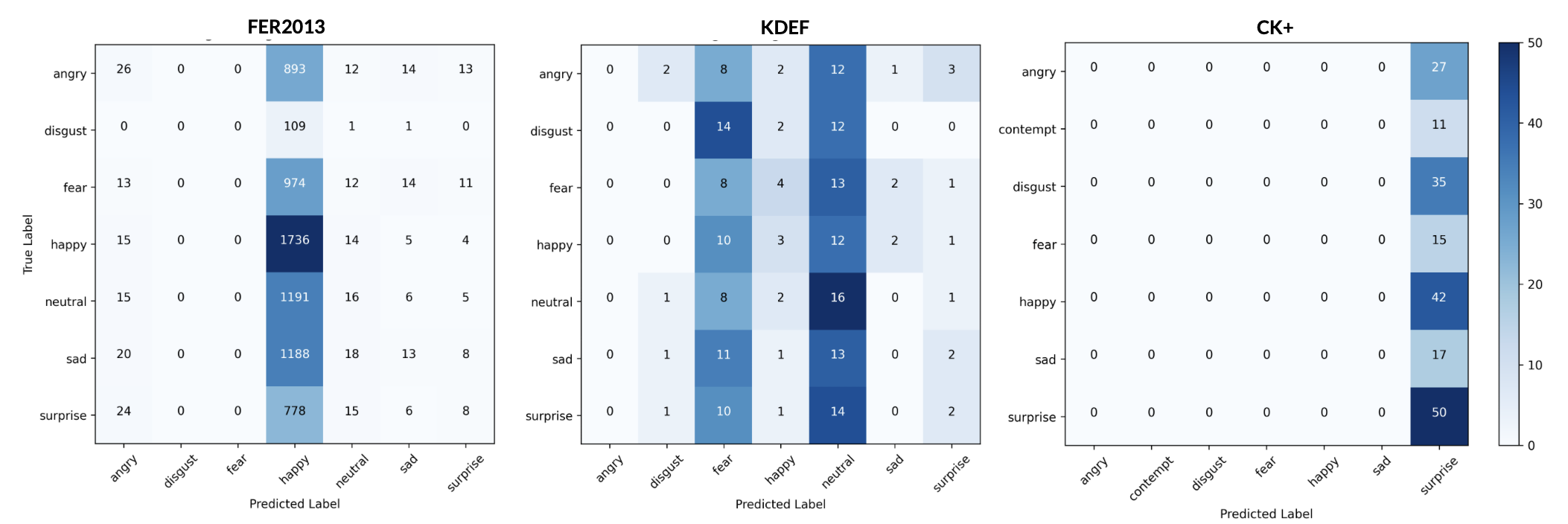}
\caption{Confusion matrices for the LBP + Logistic Regression model across FER-2013, KDEF, and CK+ datasets.}
\label{fig:lbp_confusion}
\end{figure*}

The confusion matrices shown in Figure~\ref{fig:lbp_confusion} reveal severe classification bias across multiple datasets. In FER-2013, the model predicts a dominant class for a large portion of samples, indicating poor class separability. Similar behavior is observed in CK+, where many expressions collapse into a single predicted category. This demonstrates that LBP descriptors fail to capture meaningful global facial structures required for emotion discrimination.

The poor performance can be attributed to the limitations of texture-only representations. Facial expression recognition depends heavily on spatial relationships between facial components such as the eyebrows, eyes, mouth, and cheeks. Since LBP encodes only local neighborhood texture patterns without preserving higher-level structural information, the extracted features become insufficient for robust emotion classification.

Overall, the LBP + Logistic Regression pipeline demonstrates limited effectiveness for facial expression recognition and highlights the importance of learning richer spatial representations for real-world emotion analysis tasks.

\subsection{CNN Results}

\begin{table}[htbp]
\centering
\caption{CNN Accuracy Across Datasets}
\begin{tabular}{cc}
\toprule
Dataset & Accuracy (\%) \\
\midrule
FER-2013 & 51.6 \\
KDEF & 72.4 \\
CK+ & 96.9 \\
\bottomrule
\end{tabular}
\label{tab:cnn_results}
\end{table}

The table~\ref{tab:cnn_results} summarizes the classification accuracy achieved by the CNN model across the three datasets. It is clear the CNN achieved the best overall performance across the evaluated datasets. High accuracy was obtained on both CK+ and KDEF, while the model also achieved the strongest performance on FER-2013 compared to the handcrafted feature-based approaches. These results indicate that CNNs are significantly more effective at learning robust and discriminative facial representations under varying imaging conditions.

\begin{figure}[!t]
\centering
\includegraphics[width=0.48\textwidth]{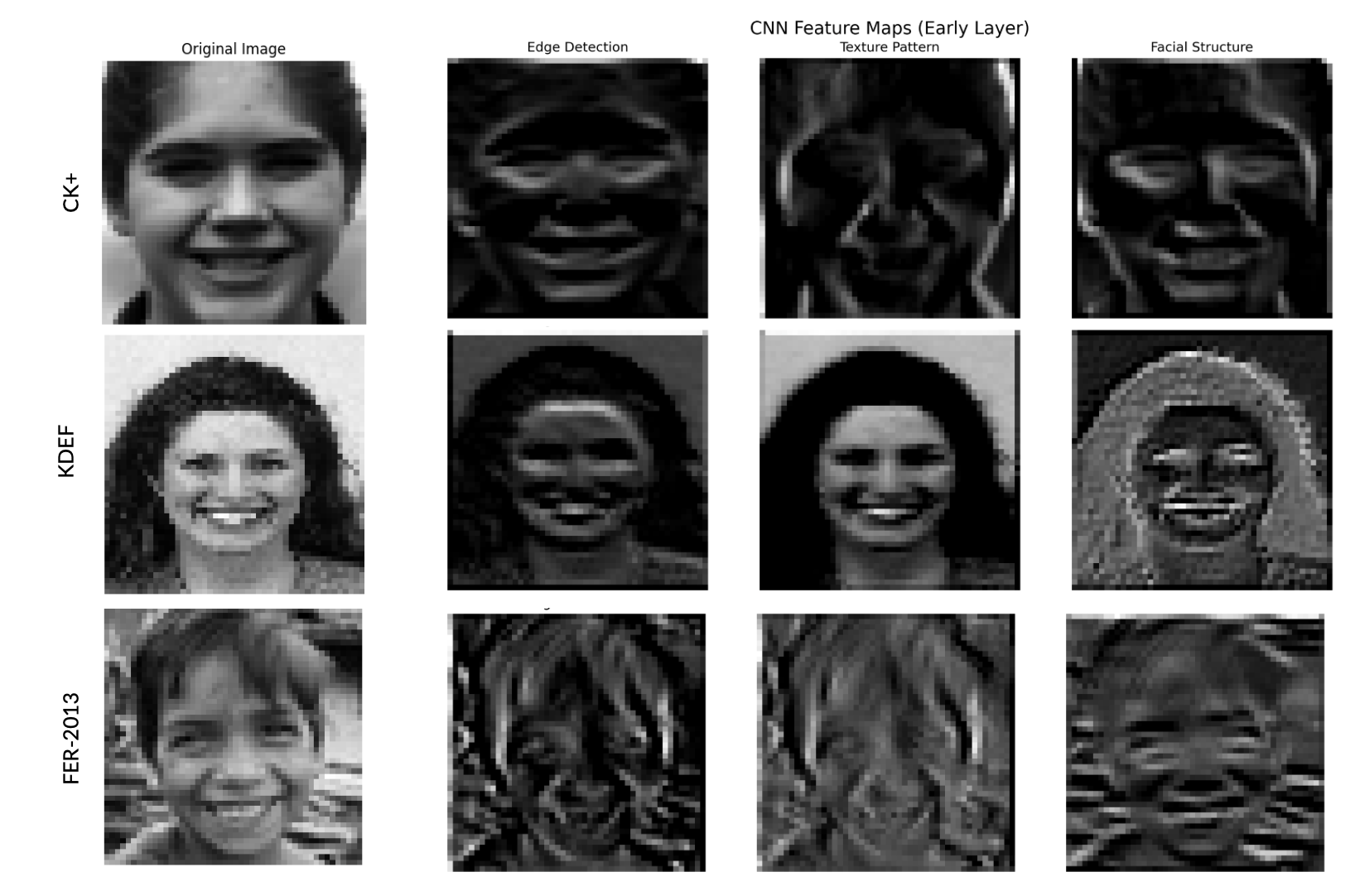}
\caption{CNN feature map visualizations from FER-2013, KDEF, and CK+ datasets. Early CNN layers capture edge information and texture patterns, while deeper activations represent higher-level facial structures relevant for emotion recognition.}
\label{fig:cnn_features}
\end{figure}

Figure~\ref{fig:cnn_features} illustrates the feature representations learned by the CNN model. Early convolutional layers respond strongly to edges and local texture patterns, similar to handcrafted descriptors such as HOG. However, deeper feature maps progressively capture more abstract facial structures related to emotion-specific regions such as the mouth, eyes, eyebrows, and cheek contours. This hierarchical feature learning enables CNNs to represent complex facial variations more effectively than handcrafted methods.

\begin{figure*}[!t]
\centering
\includegraphics[width=0.95\textwidth]{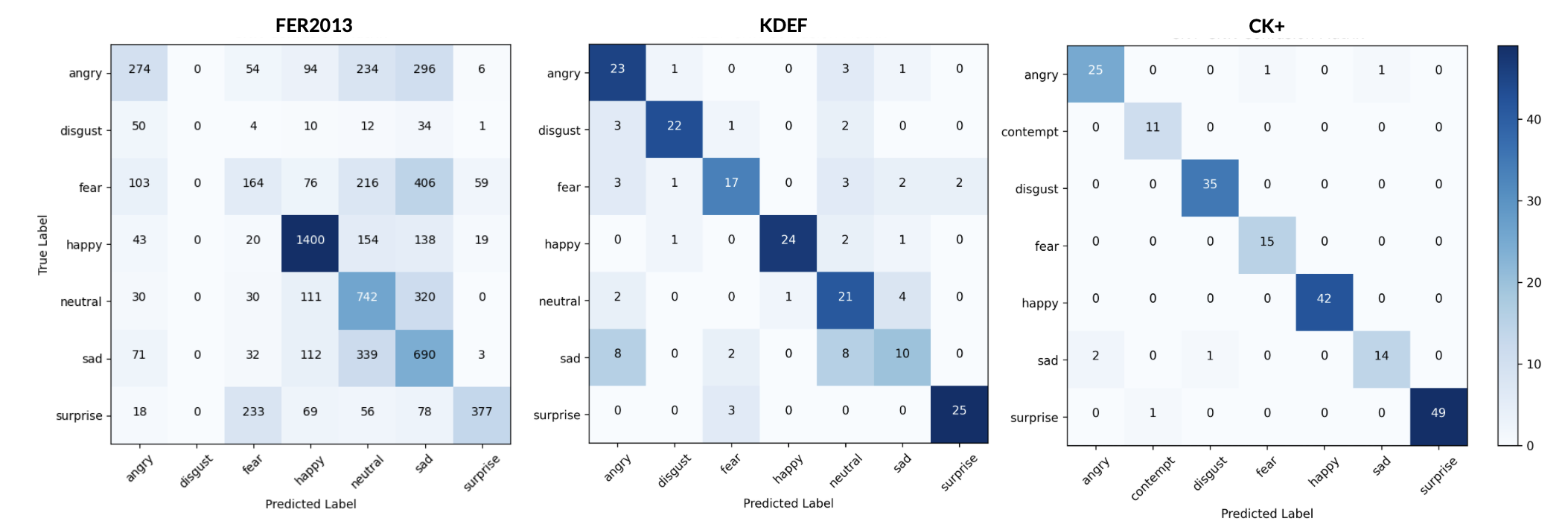}
\caption{Confusion matrices for the CNN model across FER-2013, KDEF, and CK+ datasets.}
\label{fig:cnn_confusion}
\end{figure*}

The confusion matrices shown in Figure~\ref{fig:cnn_confusion} demonstrate that the CNN achieves substantially better class separation compared to handcrafted approaches. On the CK+ dataset, nearly all emotion classes are correctly classified with minimal confusion. Similarly, KDEF shows strong class discrimination across most emotions.

Although FER-2013 remains significantly more challenging, the CNN still demonstrates improved robustness under unconstrained conditions. Some confusion persists between visually similar emotions such as fear, sadness, and neutral due to variations in pose, illumination, facial alignment, and expression intensity. Nevertheless, the CNN consistently outperforms both HOG + SVM and LBP + Logistic Regression across all datasets.

\begin{figure}[!t]
\centering
\includegraphics[width=0.49\textwidth]{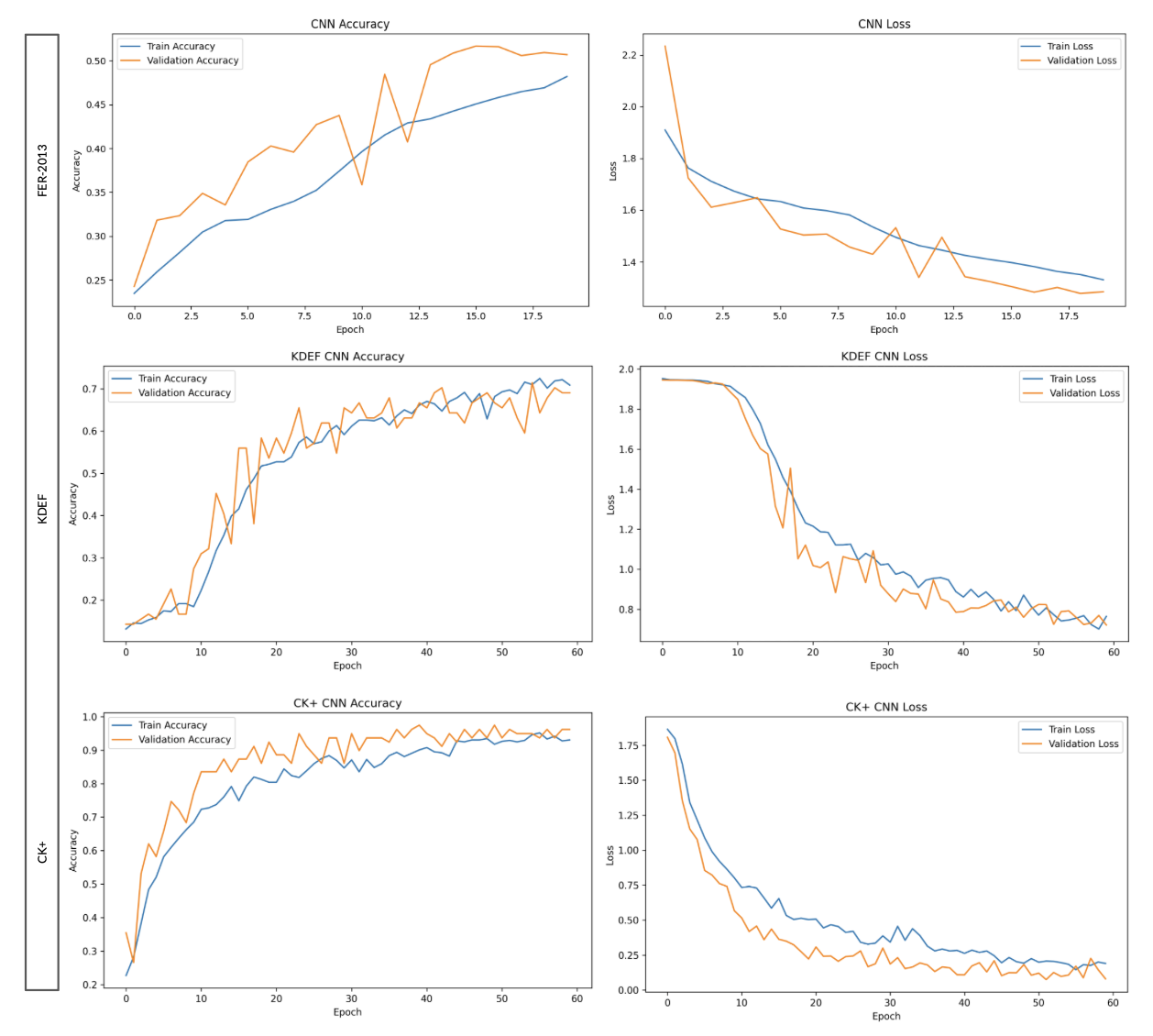}
\caption{Training and validation accuracy/loss curves for the CNN model across FER-2013, KDEF, and CK+ datasets.}
\label{fig:cnn_training_curves}
\end{figure}

Figure~\ref{fig:cnn_training_curves} illustrates the training and validation accuracy/loss curves of the CNN model across all three datasets. The CK+ dataset demonstrates the fastest convergence and highest final accuracy due to its controlled laboratory conditions and clearly distinguishable facial expressions. KDEF also shows stable convergence behavior with relatively low overfitting.

In contrast, FER-2013 exhibits slower convergence and lower final accuracy because of the large variations in illumination, pose, background clutter, and facial alignment present in unconstrained real-world images. The fluctuations observed in the FER-2013 validation curves further indicate the increased difficulty and variability of the dataset.

Overall, the training curves confirm that the CNN model generalizes effectively on controlled datasets while remaining more challenged by unconstrained real-world facial expression recognition scenarios.

The superior performance of CNNs can be attributed to their ability to automatically learn hierarchical and spatial feature representations directly from data. Unlike handcrafted methods that rely on manually designed descriptors, CNNs optimize feature extraction and classification jointly through end-to-end learning.

\subsection{Overall Comparison}

\begin{table}[htbp]
\centering
\caption{Overall Accuracy Comparison}
\begin{tabular}{cccc}
\toprule
Method & FER-2013 & CK+ & KDEF \\
\midrule
HOG + SVM & 44.0 & \textbf{98.4} & 58.2 \\
LBP + LogReg & 25.0 & 25.0 & 14.7 \\
CNN & \textbf{51.6} & 96.9 & \textbf{72.4} \\
\bottomrule
\end{tabular}
\label{tab:overall_comparison}
\end{table}

The overall comparison results in table~\ref{tab:overall_comparison} show that CNNs provide the strongest general performance across datasets, achieving the best results on FER-2013 and KDEF. HOG + SVM remains highly competitive in controlled settings, especially on CK+, where facial alignment and expression clarity favor gradient-based descriptors. However, its performance decreases as dataset complexity increases. LBP + Logistic Regression performs consistently poorly, indicating that local texture descriptors alone are insufficient for robust facial expression recognition.

\section{Discussion}

The experimental results demonstrate that dataset complexity has a significant impact on facial expression recognition performance. Both HOG + SVM and CNN achieved very high accuracy on the CK+ dataset because the images are collected under controlled laboratory conditions with aligned faces, minimal background noise, and exaggerated facial expressions. In contrast, FER-2013 proved substantially more challenging due to variations in illumination, pose, occlusion, facial alignment, and expression intensity. These factors reduce the separability between emotion classes and make robust feature extraction more difficult.

The handcrafted HOG descriptor performed strongly on controlled datasets because gradient orientation information effectively captures facial contours and edge-based structural patterns. Features around the mouth, eyes, eyebrows, and facial boundaries are highly informative for emotion classification when expressions are clearly visible. However, the performance degradation observed on FER-2013 indicates that handcrafted gradient descriptors are sensitive to real-world variability and lack strong generalization capability under unconstrained conditions.

LBP + Logistic Regression produced the weakest performance across all datasets. Although LBP successfully encodes local texture information, the resulting representations fail to capture global spatial relationships between facial components. The confusion matrices revealed strong prediction bias toward dominant classes, suggesting poor class separability. These results indicate that texture-only handcrafted representations are insufficient for reliable facial expression recognition, particularly in complex datasets.

The CNN model consistently achieved the strongest overall performance because it learns hierarchical feature representations directly from data. Early convolutional layers extract low-level edge and texture information, while deeper layers progressively capture higher-level facial structures associated with emotional expressions. Unlike handcrafted methods, CNNs optimize feature extraction and classification jointly through end-to-end learning, enabling improved robustness to illumination changes, pose variation, and noisy backgrounds.

In addition to classification performance, the compared approaches also differ in computational requirements and feature representation complexity. Handcrafted methods such as HOG + SVM and LBP + Logistic Regression are relatively lightweight because feature extraction and classification are separated into simple processing stages. These methods may be useful in resource-constrained environments where training time and computational cost are important. However, their representational capacity is limited because the extracted features are manually designed. In contrast, CNNs require more training time and computational resources, but they learn richer hierarchical representations directly from data, which improves robustness and generalization on more challenging datasets.

An important observation from this study is that very high performance on controlled datasets does not necessarily imply strong real-world generalization. While both HOG and CNN achieved near-perfect accuracy on CK+, their performance dropped noticeably on FER-2013. This highlights the importance of evaluating facial expression recognition systems on unconstrained datasets that better represent real-world environments.

Overall, the results support the hypothesis that deep learning approaches provide superior robustness and generalization for facial expression recognition tasks. Although handcrafted methods remain computationally efficient and effective in constrained settings, CNN-based approaches are more suitable for practical real-world deployment due to their ability to learn complex and discriminative facial representations automatically.

\section{Limitations}

Although the proposed study provides useful insights into the performance differences between handcrafted feature-based methods and CNNs for facial expression recognition, several limitations remain.

First, the CK+ and KDEF datasets are relatively small compared to modern large-scale facial expression datasets. The limited number of training samples may reduce model generalization capability and increase the risk of overfitting, particularly for deep learning approaches.

Second, the models were trained independently without using transfer learning or pretrained feature extractors. More advanced pretrained architectures could potentially improve feature representation quality and overall classification performance.

Third, only grayscale facial images were used in this study. While grayscale images simplify preprocessing and reduce computational complexity, important color-related information that may contribute to emotion recognition was not utilized.

In addition, the experiments focused only on static image-based facial expression recognition. Temporal facial dynamics and motion-based information, which are important for natural emotional behavior analysis, were not considered.

Finally, hyperparameter tuning and architecture optimization were limited due to computational and time constraints. More extensive tuning may further improve model performance, particularly for the CNN-based approach.

\section{Future Work}

Several directions can be explored to improve and extend this work in future studies.

First, transfer learning approaches using pretrained deep learning architectures such as ResNet~\cite{resnet}, EfficientNet, or Vision Transformers could be investigated to improve feature learning and generalization performance.

Second, hybrid approaches that combine handcrafted descriptors with CNN-based learned representations may provide complementary structural and semantic information for improved emotion recognition accuracy.

Third, future research can explore video-based facial expression recognition to capture temporal dynamics and sequential emotional transitions rather than relying solely on static images.

Additional improvements may also be achieved through stronger data augmentation techniques, larger training datasets, and cross-dataset evaluation strategies designed to assess real-world generalization capability.

Finally, explainable AI techniques such as Grad-CAM or attention visualization could be incorporated to better understand which facial regions contribute most strongly to emotion classification decisions.

\section{Conclusion}

This study presented a comparative analysis of handcrafted feature-based methods and convolutional neural networks for facial expression recognition across the FER-2013, CK+, and KDEF datasets.

The experimental results demonstrated that CNN-based approaches achieved the strongest overall performance, particularly on more complex and unconstrained datasets. HOG + SVM provided a competitive classical baseline and achieved excellent performance under controlled conditions, especially on the CK+ dataset. However, its performance decreased substantially under real-world variations such as illumination changes, pose differences, and background noise. In contrast, LBP + Logistic Regression produced consistently weak performance due to the limited representational capability of texture-only handcrafted features.

The study further showed that dataset complexity plays a major role in facial expression recognition performance. Controlled datasets such as CK+ produced very high classification accuracy, whereas FER-2013 remained significantly more challenging due to unconstrained imaging conditions and expression ambiguity.

Overall, the findings support the hypothesis that deep learning approaches are more suitable for real-world facial expression recognition tasks because they automatically learn hierarchical and spatially meaningful feature representations directly from data. Although handcrafted methods remain computationally efficient and interpretable, CNN-based approaches provide stronger robustness, scalability, and generalization capability for practical emotion recognition systems.

\balance

\bibliographystyle{IEEEtran}

\end{document}